\documentclass[
]{ceurart}

\def\year{2022}\relax
\usepackage{graphicx} 
\usepackage{natbib}  
\usepackage{caption} 
\usepackage{algorithm}
\usepackage{algorithmic}

\usepackage{hyperref}

\conference{De-Factify: Workshop on Multimodal Fact Checking and Hate Speech Detection, co-located with AAAI 2022. 2022}

\begin{document}

\copyrightyear{2022}
\copyrightclause{Copyright for this paper by its authors.
  Use permitted under Creative Commons License Attribution 4.0
  International (CC BY 4.0).}


\title{Matching Tweets With Applicable Fact-Checks Across Languages}

\author[1,3]{Ashkan Kazemi}[%
email=ashkank@umich.edu,
]
\author[1,2]{Zehua Li}[%
email=zehuali@law.stanford.edu,
]
\author[1]{Ver\'{o}nica Per\'{e}z-Rosas}[%
email=vrncapr@umich.edu,
]
\author[3,4]{Scott A. Hale}[%
email=scott@meedan.com,
]
\author[1]{Rada Mihalcea}[%
email=mihalcea@umich.edu,
]
\address[1]{Computer Science \& Engineering, University of Michigan}
\address[2]{Stanford Law School}
\address[3]{Meedan}
\address[4]{Oxford Internet Institute, University of Oxford}


\begin{abstract}
An important challenge for news fact-checking is the effective  dissemination of existing fact-checks. This in turn brings the need for reliable methods to detect previously fact-checked claims. In this paper, we focus on automatically finding existing fact-checks for claims made in social media posts (tweets). 
We conduct both classification and retrieval experiments, in monolingual (English only), multilingual (Spanish, Portuguese), and cross-lingual (Hindi-English) settings using multilingual transformer models such as XLM-RoBERTa and multilingual embeddings such as LaBSE and SBERT. We present promising results for ``match" classification (86\% average accuracy) in four language pairs. 
We also find that a BM25 baseline outperforms or is on par with state-of-the-art multilingual embedding models for the retrieval task during our monolingual experiments. We highlight and discuss NLP challenges while addressing this problem in different languages, and we introduce a novel curated dataset of fact-checks and corresponding tweets for future research. 
\end{abstract}


\maketitle

\section{Introduction}

Fact-checking is an essential part of content moderation pipelines, since it provides ground truth for veracity judgements of a given claim. 
However, manual fact-checking is slow and expensive, as it requires human expertise.




This demand has been already identified by a recent survey study \cite{nakov2021automated}, showing that fact-checkers from 24 organizations in 50 countries expressed the need for reliable methods to detect previously fact-checked claims. Recent work in natural language processing (NLP) has focused mainly on the development of automatic systems to identify misinforming claims both in monolingual \cite{shaar-etal-2020-known} and multilingual settings~\cite{kazemi-etal-2021-claim}. However, this research is mostly limited to short claims and does not directly assist the dissemination of existing fact-checks which are usually article-length documents. Moreover, existing work has addressed mainly claims in English with few exceptions such as work by~\citet{kazemi-etal-2021-claim} and the CheckThat! Lab \cite{CheckThat:ECIR2021}, an evaluation lab that included a claim retrieval challenge in English and Arabic in their 2021 edition.  


To contribute 
to this research direction, 
our paper addresses the problem of matching and finding applicable fact-checks to social media posts. 
We approach this problem using two strategies (i) fact-check ``matching'', i.e., determining whether a social media post (tweet) and a fact-check pair match or not, and (ii) fact-check ``retrieval'', i.e., given a social media post (tweet), rank and return the most relevant fact-checks discussing the claims made in it. We address the ``matching'' task by building a binary classifier on top of XLM-RoBERTa (XLM-R), a large transformer-based multilingual language model~\cite{conneau-etal-2020-unsupervised}. For the ``retrieval'' task, we build an embedding similarity search system using sentence embeddings from LaBSE \cite{feng2020language}, SBERT \cite{reimers-gurevych-2019-sentence} models and pairwise cosine similarity.
We also analyze these tasks for languages other than English, i.e., Spanish and Portuguese. Further, we investigate a cross-lingual scenario where we seek to identify applicable English fact-checks to Hindi tweets. 




\begin{table*}[t]
\caption{An example tweet and a matching fact-check (both in English) from our dataset. The fact-checking article is redacted and can be found at \href{https://www.boomlive.in/kerala-floods-fake-news-about-dam-burst-no-power-create-panic/}{this URL}.}

\begin{tabular}{| p{\columnwidth - 12pt} |}
\hline
\textbf{Tweet \#1:} Heartbreaking to see a barrage of fake WhatsApp forwards, kooky safety instructions, hysteria that \textbf{dams are breaking}, transformers are submerged and \textbf{electricity is being cut off}, hindering rescue efforts in Kerala. \\
\hline
\textbf{Tweet \#2:} Heard a news that a \textbf{shutter of Cheerakuzhy dam, Thrissur broke.} Can someone pls confirm this news. Yet to find any reference in MSM. 
If true, pls inform authorities immediately. \#KeralaFloods2018 \\
\hline
\textbf{Fact-Check Report:} The Kerala government already on the back foot trying to battle a massive crisis due to relentless rain and flooding over the past week now have one more big worry - fake news led by incorrect reporting and rumours. The floods have already resulted in the deaths of 324 people in the past 17 days with thousands of people stranded across the state on rooftops and relief camps. ... [redacted]

Worst of all is an audio clip in which a person is heard saying that the Mullaperiyar \textbf{dam has developed cracks} and in the next three hours, the downstream districts of Idukki, Ernakulam, Thrissur and Allapuzha will be washed away. He urges people to take it seriously as the government is hiding the information about the leak and that he got to know of it from a friend who works in Modi's office. (We have not uploaded the audio clip in the story to prevent further panic). ... [redacted]

Yet another audio message was going around claiming that the \textbf{shutters of Cheerakuzhi dam} built across Gayatri river (also know as Bharathpuzha) in Thrissur \textbf{are damaged.} However, regional media Manorama News and Deshabhimani clarified that it is not true and this exaggerated message is meant to create a scare. ... [redacted]

Another message which has created quite a scare is that the \textbf{whole state will have no electricity} tomorrow as the Kerala State Electricity Board (KSEB) will shut down its operations. [redacted] However, KSEB and Kerala Police were quick to respond and call the message fake. KSEB clarified through a Facebook post that KSEB employees are engaged in relentless efforts to restore electricity in the areas facing power cuts. To avoid danger during floods, power supply and production in certain parts have been temporarily discontinued. However, as the water recedes the power supply will be restored. The electricity board also appealed to people to not spread these rumours. ... [redacted] \\
\hline
\end{tabular}
\label{data_example}
\end{table*}

\section{Related Work}
While fully automatic fake news detection and fact-checking systems \cite{perez-rosas-etal-2018-automatic, thorne-vlachos-2018-automated} remain an active research topic within the NLP community, there have been new research fronts in the fight against misinformation, including claim matching \cite{shaar-etal-2020-known, kazemi-etal-2021-claim}, check-worthiness detection \cite{hassan2017toward, konstantinovskiy2021toward}, explanations \cite{kazemi-etal-2021-extractive, atanasova-etal-2020-generating, kotonya-toni-2020-explainable}, and detecting out of context misinformation \cite{da-etal-2021-edited, aneja2021cosmos}.

On the context of claim matching, \citet{shaar-etal-2020-known} introduced a retrieval-based version of the task where, for a given input claim, the goal is to rank similar check-worthy claims based on their relevance to the input claim. For this task, they focus on political related claims in English and a presented a rank model that relied on BERT~\cite{devlin-etal-2019-bert} and BM25 based architectures. More recently, \citet{kazemi-etal-2021-claim} focused on matching claims that can be served with one fact-check in five low and high-resource languages. Similarly \citet{vo-lee-2020-facts} conducted claim matching in a multimodal setting where they find previously debunked texts and images. In addition, The CheckThat! Lab \citeyear{CheckThat:ECIR2021} evaluation presented claim matching as a shared task for English and Arabic.

Although works such as \citet{shaar-etal-2020-known} have matched English tweets and fact-checks, most of prior work has mainly focused on matching claims with other similar claims that are usually short in length. In this paper, we seek to match claims with applicable fact-check reports that are significantly longer and potentially express the claim in different ways. Additionally, we approach the claim matching problem in multilingual and cross-lingual settings and experiment with recent neural models in multilingual NLP.

Among them, we use XLM-RoBERTa \cite{conneau-etal-2020-unsupervised}, a powerful multilingual transformer-based language model that have achieved competitive performance on cross-lingual and multilingual benchmarks. The model is trained on more than 2TBs CommonCrawl data and supports one hundred languages. We also rely on recent language agnostic embedding models such as LaBSE (Language-agnostic BERT Sentence Embedding) \cite{feng2020language}, a sentence embedding model that can produce embeddings in 109 languages. This model was built using a combined pretraining method of masked and translation language modeling trained on 17 billion monolingual sentences from CommonCrawl and 6 billion translated pairs of sentences. Sentence-BERT (SBERT) \cite{reimers-gurevych-2019-sentence} use twin and triplet networks on top of language models for producing sentence embeddings. In their follow up work to SBERT \cite{reimers-gurevych-2020-making}, they also propose an approach to convert monolingual embeddings into multilingual ones. We also use Elasticsearch's implementation of the BM25 retrieval system \cite{robertson2009probabilistic}, which provides fast and scalable text search.

\section{Data}
Our data is derived from 150,000 fact-checks obtained from several sources, including (i) fact-checking organizations certified by the International Fact-Checking Network (IFCN) and (ii) fact-checking aggregators such as Google Fact-check Explorer,\footnote{\url{https://toolbox.google.com/factcheck/explorer}} GESIS~\cite{tchechmedjiev2019claimskg}, and Data Commons.\footnote{\url{https://datacommons.org/factcheck/faq}} The collected fact-checks cover several languages, including English, Spanish, Portuguese, and Hindi. Each fact-check includes a claim and usually a justification article for the claim verdict, and metadata such as publication date, claim veracity and references to the original content that needed the fact-check.

\begin{table*}
  \caption{Per language statistics of our (tweet, fact-check) dataset.}
  \label{table:dataset}
  \begin{tabular}{ c c c }
    \toprule
    Tweet Language & Article Language & \# of Pairs \\ 
    \midrule
    English & English & 4,850 \\ 
    Hindi & English & 664 \\
    Spanish & Spanish & 617 \\  
    Portuguese & Portuguese & 402 \\
  \bottomrule
\end{tabular}
\end{table*}

Similar to \citet{shahi2020amused} and \citet{shahi2021exploratory}, we use social media links included in the fact-checks and their original news sources (whenever available) to build a dataset consisting of (tweet, fact-check) pairs. Given that the fact-checks include several languages, we obtain monolingual pairs in English, Spanish, and Portuguese and also cross-lingual pairs consisting of Hindi tweets and English fact-checks. In cases where the tweet contains a link (usually to a news article), we also append the preview text from the link to the tweet text, to capture more of the tweet's context.

Since we match tweets and fact-checks automatically through references in the text we conducted an additional verification step to make sure that the identified pairs are indeed related. 
We thus annotate a random sample of 100 English (tweet, fact-check) pairs to verify whether each fact-check is applicable to its matched tweet. The annotation was conducted independently by two annotators, reaching an 87\% agreement between annotator responses. We find that 89\% of the tweets in our sample matched their corresponding fact-checks and in most cases the pairs include at least one fact-check worthy claim. This finding suggests that while there is some degree of noise in the pairing process, most of the pairs are correct matches. Table \ref{table:dataset} shows a summary of the final set of (fact-check,tweet) pairs per language. Sample (fact-check,tweet) pairs are shown in Table \ref{data_example}. Note that multiple tweets can be matched to the same fact-check. 



\section{Models \& Baselines}

\subsection{Matching (tweet, fact-check) Pairs}

We address the task of matching (tweet, fact-check) pairs as a binary classification problem using ``match'' or ``not match'' as possible labels. 


Our dataset consists of only positive labels  since we only collected matching (tweet, fact-check) pairs, and training a binary classifier also requires negative examples, so we explored several strategies to obtain negative samples. Initially, we selected negative examples by randomly pairing non-matching tweets and fact-checks. We then built a binary classifier using an XLM-R model fine-tuned on the resulting dataset. However, preliminary evaluations showed that the resulting classifier was not able to generalize well. 
We believe this is due to the classifier's lack of exposure to challenging negative samples, since most of random pairings are easily distinguished from matching (tweet, fact-check) pairs.

In order to get more challenging negative samples, we opted for finding non-matching (tweet, fact-check) pairs based on their pairwise similarity. We start by calculating the pair-wise cosine similarity across all possible (tweet, fact-check) pairs in the dataset, within the same multi/cross-lingual setting. Then, we use LaBSE embeddings \cite{feng2020language} of tweets and fact-check articles and rank non-matching pairs by decreasing cosine similarities. Next, we pick the top negative samples from this set, i.e., pairs with similarities lower than 0.7, to reduce the number of false negatives.
We train our XLM-R classifier with the resulting data and find an 15\% absolute improvement of classification accuracy as compared to training on randomly selected pairs.%

Since our dataset contains multiple languages, we conduct an additional set of experiments where we train separate classifiers for each language pair e.g., English, Spanish, Portuguese, Hindi-English, as well as a classifier that uses pairs in all languages. Results are presented in Table~\ref{table:classifier}.

\begin{table*}[t]
\caption{Results from matching (tweet, fact-check) pairs as a binary classification problem. F1+ and F1- refer to the F1 score for the ``match'' and ``not match'' classes.}
\centering
\begin{tabular}{ l c c c c c c } 
 \toprule
  & \multicolumn{3}{c}{\textbf{Trained Separately}} &  \multicolumn{3}{c}{\textbf{Trained Altogether}} \\
 \textbf{Lang. Pairs} & Acc. & F1+ & F1- & Acc. & F1+ & F1- \\
 \midrule
En-En & 88.46\% & 88.72\% & 88.17\% & 88.61\% & 88.66\% & 88.54\% \\
Hi-En & 80.27\% & 80.53\% & 79.71\% & 80.50\% & 81.60\% & 78.90\% \\
Es-Es & 85.82\% & 86.07\% & 85.09\% & 88.57\% & 88.93\% & 88.06\% \\
Pt-Pt & 84.08\% & 83.67\% & 83.59\% & 87.44\% & 87.65\% & 87.25\% \\
 \bottomrule
\end{tabular}
\label{table:classifier}
\end{table*}


\subsection{Finding Applicable Fact-Checks for Tweets}

A different perspective on the problem of matching fact-checks with tweets is to retrieve and rank fact-checks based on their relevance to an input tweet. As opposed to binary classification, this approach provides a ranked list of options to choose from and requires human intervention to select the most appropriate fact-check. This strategy makes the search process more scalable since finding applicable fact-checks does not require the quadratic number of computationally expensive comparisons that make the binary classification approach computationally intractable for retrieval. 

During our experiments we use BM25 as our baseline retrieval method. We use the implementation provided in Elasticsearch~\cite{robertson2009probabilistic} . 
BM25 is inherently language agnostic since it relies on token matching. However, this makes it unable to handle cross-lingual text, which is the case of our set of (Hindi tweets, English fact-checks). To address this issue, we translate the Hindi tweets into English using Google translate before using BM25. Our preliminary experiments show that the use of translated tweets leads to a stronger baseline as compared to just applying BM25 to the original Hindi tweets. The translation is only to accommodate for the lack of cross-lingual operability of BM25 and is necessary for keeping consistent with our comparison methodology.

Additionally, we experiment with multilingual sentence embeddings, namely LaBSE and (multilingual) MPNet-SBERT. 
Since these embedding models only support inputs up to 512 tokens and fact-check articles are usually longer, we embed article paragraphs instead of whole articles. Thus, we compare an input tweet with paragraphs from the fact-check reports and not with full-length articles. Note that unlike the embedding-based models, BM25 is able to handle text in arbitrary length, so in order to carry out a fair comparison of the baseline and embeddings, we additionally provide a BM25 baseline using article paragraphs only.


The results for these experiments are presented in tables \ref{table:english_results}, \ref{table:other_language_results} and \ref{table:crosslingual_results}. We discuss them in detail in the next sections of the paper.

\subsection{Experimental Setup}
We use HuggingFace's \textit{transformers} \cite{wolf-etal-2020-transformers} and the SBERT library to implement our models. We run our code on a GPU-enabled server. For the English retrieval experiments, we use LaBSE and the \textit{paraphrase-mpnet-base-v2} model which we call ``MPNet-SBERT'', an SBERT embedding model trained on top of MPNet \cite{song2020mpnet}. We use the multilingual version of the same model (\textit{paraphrase-multilingual-mpnet-base-v2}) for Spanish, Portuguese and Hindi-English pairs.

To evaluate classification tasks, we use accuracy and F1 score as our main metrics. The classification experiments are conducted using 5-fold cross validation. For our retrieval experiments, we use ``mean reciprocal rank'' (MRR) and ``mean average precision'' (MAP@K) for different values of K.

\section{Experiments in English}


Results in Table~\ref{table:classifier} show a promising performance from our XLM-R models in matching tweet-fact check pairs in English, with accuracies of up to 89\%. As observed, there is a slight performance increase when training the model with all languages as compared to using English only. While the increase in performance when training altogether is more significant for other languages, it is worth noting that the English performance remains robust to noise as using training data from other languages can introduce noise for a model applied on English only. Although there is a slight performance decrease in the ``match'' class, the performance gain for the ``not match" class when using the training altogether model is large enough to improve the overall accuracy, which suggests potential benefits from using data in other languages.


Table \ref{table:english_results} presents results for the retrieval-based evaluation. The full-length BM25 baseline achieves 65\% MAP@1 and 72\% MRR scores as the best performing model.   
The gap between MAP numbers mostly decreases as K increases which is an expected behavior for mean average precision. At first glance, it seems that feeding paragraphs from the article to the embedding models could account for the performance loss, since the full-length BM25 uses the whole document at once, therefore providing the upper bound performance for this task.
While the paragraph BM25 system has a decrease in performance relative to full-length BM25, not all of the performance gap between embedding models and BM25 can be explained by the inability of embedding models to process longer documents. Among the embedding based models, MPNet-SBERT is the best performing model achieving 54\% MAP@1 and 62\% MRR. The second best performing model, LaBSE, is behind MPNet-SBERT by a noticeable margin of more than 8 MAP@1 and MRR points. A potential explanation for this performance decrease is that LaBSE is a multilingual model and performance decrease with respect to single-language models is often observed when a model supports multiple languages (100+ in LaBSE's case) at once.

Even though we see promising classification results, our experiments show that state-of-the-art NLP algorithms are still unable to compete against the BM25 baseline in finding applicable fact-checks.

\begin{table*}[ht]
\caption{Results from retrieval experiments in English.} 
\centering
\begin{tabular}{ l c c c c c c } 
 \toprule
  & \multicolumn{5}{c}{\textbf{MAP@K}} & \textbf{MRR} \\
 \textbf{Model} & K=1 & K=5 & K=10 & K=20 & K=50 & \\
 \midrule
 Full-Length BM25 & 64.85\% & 70.82\% & 71.18\% & 71.30\% & 71.39\% & 71.51\% \\
Paragraph BM25 & 62.03\% & 66.68\% & 67.27\% & 67.46\% & 67.57\% & 67.61\% \\
LaBSE & 44.98\% & 51.59\% & 52.24\% & 52.64\% & 52.81\% & 53.00\% \\
MPNet-SBERT & 53.56\% & 60.58\% & 61.20\% & 61.48\% & 61.68\% & 61.84\% \\
 \bottomrule
\end{tabular}
\label{table:english_results}
\end{table*}

\section{Experiments in Other Languages}\label{section:nonenglish}
Since our dataset also covers Spanish and Portuguese, we conduct an additional set of experiments to assess the performance of our models in languages other than English. During these experiments, we test the same models used with English, with the exception of English MPNet-SBERT that was replaced with the multilingual version. 

The results in Table \ref{table:classifier} indicate that training a single XLM-R model on data from all languages performs more accurately on average (86.28\%) in comparison with training separate models per language (84.66\%) for matching. Particularly, we see a performance increase for Spanish and Portuguese, with accuracies of up to 88.57\% and 87.44\% respectively.  
Training a single XLM-R model on all languages leads to a performance improvement up to 3.36\% for Spanish and Portuguese as compared to the single-language models, implying the transfer of task expertise across languages for XLM-R. A potential explanation of the fact that a single model has the leverage of larger data. We believe this is particularly effective when the languages are similar and can learn from each other's data. Also note that classifying (tweet, fact-check) pairs in multiple languages with a single XLM-R model is preferred since it saves computational resources and is easier to use.


We observe mostly similar trends to the English experiements for fact-check retrieval as shown in Table \ref{table:other_language_results}, with two exceptions: (i) multilingual MPNet-SBERT slightly outperforming the paragraph BM25 model by 1.19 MAP@1 and 2.55 MRR points in Spanish and (ii) LaBSE outperforming multilingual MPNet-SBERT by 5 MAP@1 and 2 MRR@1 points for Portuguese. Even though LaBSE performed worse than MPNet-SBERT in Spanish (2\% MAP@1 and MRR), together with the Portuguese results, LaBSE would still be the preferred embedding in non-English languages such as Spanish and Portuguese.

Note that during these experiments, the embedding models mostly underperformed in comparison to both BM25 baselines, with the full-length BM25 outperforming the best embedding model by 14 MAP@1 and 11 MRR points in Spanish in comparison with MPNet-SBERT and 10 MAP@1 and MRR points in Portuguese in comparison with LaBSE.

\begin{table*}[ht]
\caption{Results from retrieval experiments in Spanish and Portuguese. ML in ``ML MPNet-SBERT'' is short for multilingual.}
\centering
\begin{tabular}{ l c c c c c c } 
 \toprule
 \textbf{Model} & \multicolumn{5}{c}{\textbf{MAP@K}} & \textbf{MRR} \\
 \midrule
 \textbf{Spanish} & K=1 & K=5 & K=10 & K=20 & K=50 &  \\
 \midrule
Full-Length BM25 & 73.41\% & 78.56\% & 78.78\% & 78.84\% & 78.90\% & 78.54\% \\
Paragraph BM25 & 58.33\% & 63.65\% & 64.38\% & 64.78\% & 64.88\% & 64.56\% \\
LaBSE & 57.14\% & 62.83\% & 63.68\% & 64.01\% & 64.24\% & 64.68\% \\
ML MPNet-SBERT & 59.52\% & 66.28\% & 66.58\% & 66.74\% & 66.90\% & 67.23\% \\
 \midrule
 \multicolumn{7}{l}{\textbf{Portuguese}} \\
 \midrule
Full-Length BM25 & 69.62\% & 74.09\% & 74.73\% & 74.99\% & 75.04\% & 75.04\% \\
Paragraph BM25 & 69.62\% & 72.51\% & 72.97\% & 73.23\% & 73.32\% & 73.32\% \\
LaBSE & 59.49\% & 62.95\% & 63.25\% & 63.53\% & 63.89\% & 63.89\% \\
ML MPNet-SBERT & 54.43\% & 60.06\% & 61.07\% & 61.29\% & 61.55\% & 61.55\% \\
 \bottomrule
\end{tabular}
\label{table:other_language_results}
\end{table*}

\section{Cross-Language Experiments}
The retrieval results are presented in Table \ref{table:crosslingual_results}. Unlike the monolingual experiments, models from the previous section outperform BM25 by noticeable margins in the retrieval setting and perform competitively with other language pairs in classification too. The only difference is that for the cross-lingual Hindi-English pairs in retrieval, the tweets are first translated into English. Also, the single XLM-R model trained on data from all language pairs classifies (Hindi tweet, English fact-check) pairs comparably with the monolingual models with 80.57\% accuracy according to Table \ref{table:classifier}. Although there is a 5.8\% accuracy decrease compared to the best mean accuracy (altogether), the Hindi-English XLM-R performance is still considered competitive for the more difficult task of cross-lingual matching.

Furthermore, we observe high cross-lingual performance from LaBSE, better than its performance on English and close with Portuguese and Spanish. LaBSE outperforms the best BM25 system (full-length BM25) by about 7.5 MAP@1 and 7 MRR points. However, there is a large performance gap between the embedding models (12\% MAP@1, 9.5\% MRR) as multilingual MPNet-SBERT has the worst performance of all systems but still performs not too far worse than the BM25 baselines. The improvement of LaBSE over ML MPNet-SBERT can be attributed to the fact that LaBSE was trained specifically for cross-lingual representation learning. We believe that BM25's underperformance can be attributed to translation errors. However,  this is one of the few ways (if not the only way) that BM25 can support cross-lingual queries, ultimately making this a downside of using BM25. Overall, the use of XLM-R and LaBSE for cross-lingual matching and retrieval of fact-checks is a promising direction.

\begin{table*}[ht]
\caption{Results from cross-lingual retrieval experiments with tweets in Hindi and fact-check articles in English. For BM25 systems, the tweet is first translated into English before being fed to BM25.}
\centering
\begin{tabular}{ l c c c c c c } 
 \toprule
  & \multicolumn{5}{c}{\textbf{MAP@K}} & \textbf{MRR} \\
 \textbf{Model} & K=1 & K=5 & K=10 & K=20 & K=50 & \\
 \midrule
Full-Length BM25 & 47.95\% & 52.59\% & 52.99\% & 53.11\% & 53.15\% & 52.80\% \\
Paragraph BM25 & 45.89\% & 50.31\% & 50.78\% & 50.98\% & 51.02\% & 50.48\% \\
LaBSE & 55.48\% & 59.12\% & 59.63\% & 59.92\% & 60.14\% & 59.79\% \\
ML MPNet-SBERT & 43.15\% & 48.72\% & 49.17\% & 49.68\% & 49.87\% & 50.22\% \\
 \bottomrule
\end{tabular}
\label{table:crosslingual_results}
\end{table*}

\section{Discussion and Future Work}

Our experiments show promising performance from XLM-R in the matching classification of tweets and fact-check pairs, with the single XLM-R model trained on all data performing on average 86.28\% accurately. While the binary XLM-R classifier performs reasonably well on full-length articles, we found that it does not perform as well in classifying (tweet, fact-check paragraph) pairs when we used it to refine retrieval results. Reranking classifiers have shown promising results in prior work \cite{nogueira2019passage}, however they were not particularly applicable in our case since we do not have paragraph-level labels for the fact-check articles to train a classifier that can rerank paragraphs.

Both BM25 baselines outperform or perform competitively with state-of-the-art multilingual sentence embedding models in monolingual retrieval settings. However, there is a key difference between BM25 and the embedding models: BM25 can handle articles of arbitrary length, whereas both LaBSE and MPNet-SBERT can handle only up to 512 tokens of input. This is a source of performance loss for LaBSE and SBERT in our task. In future work, we plan to explore long document transformers such as Longformer \cite{beltagy2020longformer} to address this problem. We believe that a long document multilingual embedding model can provide improvements not only to our research problem, but to many other areas such as news NLP and legal document processing in multilingual settings.

It is important to note that the input length limit does not explain all of the performance gap. We found that the paragraph BM25 system still outperforms the embedding-based systems by at least 10 MAP@1 and 9.5 MRR points in Portuguese experiments and performs similarly to LaBSE and MPNet SBERT in Spanish. While specialized embeddings like question answering embedding models exist for English through SBERT, they are neither necessarily applicable in searching through fact-checks nor easy to come by in non-English, multilingual and cross-lingual capabilities. Building specialized embedding models for searching through applicable fact-checks is a promising next step in improving the embedding-based retrieval systems.


LaBSE provides impressive results on cross-lingual (Hindi tweet, English fact-check) pairs, outperforming BM25 baselines and MPNet-SBERT as a single multilingual embedding model with support for more than one hundred languages. This is an important problem to solve, since misinformation travels across borders and being able to search through fact-checks across different languages can save a great deal of manual fact-checking efforts. Therefore, cross-lingual search of applicable fact-checks for social media posts has a great potential for extending the reach of manual fact-checking. Furthermore, since LaBSE performs better or comparable with multilingual MPNet-SBERT overall, this makes it the better choice for embedding models when searching for relevant fact-checks on non-English social media.

\section{Conclusion}
In this paper, we approached a new version of the ``claim matching'' problem in which we match applicable fact-checks with social media posts to increase the reach of manual fact-checking. We addressed this problem using classification and retrieval based strategies in multiple languages (English, Hindi, Portuguese and Spanish). We provided new benchmarks for this task, which we will release publicly upon the paper's acceptance.

Our results showed promising performance as 
we are able to classify matching (tweet, fact-check) pairs with accuracies of up to 89\% in four language pairs. 
From our retrieval experiments we found that monolingual pairs of (tweet, fact-check) are better retrieved by BM25, which meaningfully outperforms state-of-the-art multilingual embedding models in the retrieval task. Despite this, we observe promising performance in cross-lingual settings with LaBSE achieving more than 7.5 MAP@1 points improvement over the best BM25 baseline. 

We identified the monolingual retrieval of applicable fact-checks as a challenging area for state-of-the-art NLP and highlighted the need for specialized and long document multilingual embeddings as an important direction for future work.

Our newly curated multi/cross-lingual dataset of (tweet, fact-check) pairs in English, Spanish, Portuguese and Hindi is publicly available at \href{http://lit.eecs.umich.edu}{http://lit.eecs.umich.edu}. 

\bibliography{anthology, aaai22}

\end{document}